\documentclass[runningheads]{llncs}
\usepackage[T1]{fontenc}
\usepackage{graphicx,verbatim}
\usepackage{amsmath}
\usepackage{amsfonts}
\usepackage{bm}
\usepackage{bbm}
\usepackage{booktabs}
\usepackage{multirow}
\usepackage{makecell}
\usepackage{hyperref}
 \hypersetup{
     colorlinks=true, 
     citecolor=green  
 }
\usepackage{color}
\usepackage{xcolor}

\newcommand{\inlinesec}[1]{{\vspace{0.5em}\noindent\textbf{#1}}\hspace{0.5em}}

\newcommand{\ours}{\textsc{STRIQ}}

\begin{document}

\title{Subspace-Guided Semantic and Topological Invariant Registration for Annotation-Free Ultrasound Plane Quality Control}
\titlerunning{Subspace-Guided Registration for Ultrasound Quality Control}


\author{
  Chunzheng Zhu\inst{1}
  \and Jianxin Lin\inst{1}\thanks{Corresponding authors.}
  \and Feng Wang\inst{1}
  \and Cheng Jiang\inst{1}
  \and Guanghua Tan\inst{1}$^{\star}$
  \and Zhenyu Zhou\inst{1}
  \and Shengli Li\inst{2}
  \and Kenli Li\inst{1}
}

\authorrunning{C. Zhu et al.}
\titlerunning{Subspace-Guided Registration for Ultrasound Quality Control}
\institute{Hunan University, Changsha, China\\
\email{\{linjianxin, guanghuatan\}@hnu.edu.cn}
\and
Shenzhen Maternity and Child Healthcare Hospital, Shenzhen, China\\
}
\maketitle

\begin{abstract}
Reliable quality control (QC) of ultrasound images is essential for both real-time acquisition guidance and retrospective clinical audit, yet existing approaches rely heavily on per-plane annotations, or employ pseudo-labeling prone to systematic bias under spatial deformations inherent in clinical acquisition. We present \textbf{\ours{}}, a registration-driven framework that recasts annotation-free US plane quality control as a \emph{subspace-guided consistency measurement} problem. Specifically, \ours{} introduces a \textit{\textbf{Latent Registration Aligner}} (LRA) to establish hierarchical feature space correspondences between query images and variance-driven anchors, which are autonomously distilled from unlabeled data via a variance spectrum criterion to serve as structurally stable prototypes. To further disambiguate anatomical planes and mitigate negative knowledge transfer, we propose an \textit{\textbf{Orthogonal Knowledge Subspace}} (OKS) module. The OKS decomposes plane-specific representations into mutually orthogonal subspaces, enabling fine-grained expert collaboration while preventing inter-plane interference, ensuring that the quality metric is grounded in principled subspace proximity. Extensive experiments on the in-house US4QA and public CAMUS datasets demonstrate that \ours{} achieves state-of-the-art correlation with clinical quality scores, establishing a new paradigm for annotation-free, real-time reliable ultrasound quality control. Our code is available at \url{https://github.com/zhcz328/STRIQ}.
\end{abstract}

\section{Introduction}

\begin{figure}[t]
\centering \includegraphics[width=1\linewidth]{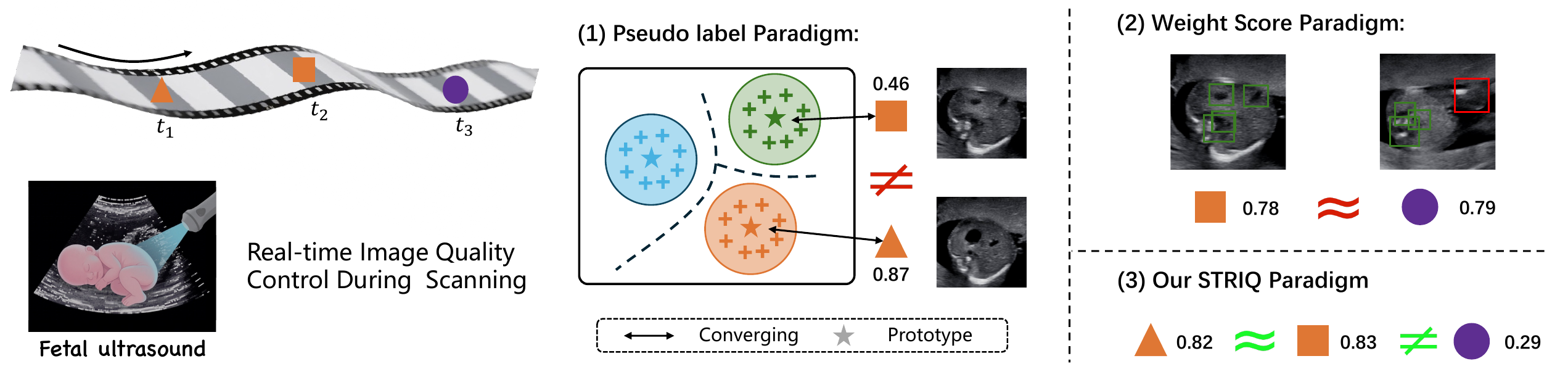} 
\caption{Ultrasound QC paradigm comparison. 
\ours{} addresses the \text{instability} of pseudo-labeling under complex deformations; and  overcomes the structure-weighted detection errors prevalent in annotation-heavy weighting schemes. 
} 
\vspace{-4mm}
\label{fig:motivation} 
\end{figure}

Ultrasound imaging is the primary modality for prenatal screening and diagnosis, yet its fidelity is highly dependent on operator expertise, probe positioning, and acoustic conditions~\cite{zhu2024advancing,jiang2025mcbl}. Suboptimal image quality can elevate misdiagnosis rates by up to 30\% in certain clinical applications~\cite{dudley2010guidelines}, a risk amplified in fetal anomaly screening where subtle degradations may obscure early-stage abnormalities~\cite{carvalho2013isuog,karim2025impact}. Clinical quality control (QC) operates along two temporal axes: \textit{intra-procedural} QC for real-time feedback during acquisition, and \textit{post-procedural} QC for retrospective compliance auditing. Both modalities demand automated, annotation-free solutions capable of accommodating the geometric variability inherent in clinical practice.


Existing automated approaches fall into three paradigms. As illustrated in Fig.~\ref{fig:motivation}, \textit{pseudo-label approaches}~\cite{guo2024unsupervised,raina2023expert} generate surrogate labels via feature-space clustering, but remain sensitive to spatial deformations ubiquitous in clinical ultrasound, yielding biased results under morphological variation~\cite{balakrishnan2019voxelmorph}. \textit{Structure-detection methods}~\cite{wu2017fuiqa,dong2019generic,lin2019multi} infer quality from anatomical landmark confidence, yet overlook perceptual context and demand extensive annotations. \textit{General-purpose image quality assessment (IQA) methods}~\cite{mittal2012no,agnolucci2024arniqa} transfer poorly due to the domain gap between natural and medical images~\cite{miao2025ultrasound}. Critically, none explicitly model the geometric transformations underlying quality degradation, nor leverage topological invariance as an assessment prior.

Recent advances in feature-level image registration have demonstrated that learned representations can establish robust anatomical correspondences while accommodating appearance variations~\cite{czolbe2023semantic,chen2025survey,liu2025survey}. This suggests a principled reformulation: the \textit{registrability} of a query image to high-quality reference standards serves as a natural proxy for image quality. However, when a single registration network is shared across heterogeneous anatomical planes with distinct structural topology, the model encounters \textit{negative transfer}, wherein conflicting plane-specific knowledge degrades overall performance.

\begin{figure}[t]
\centering
\includegraphics[width=\textwidth]{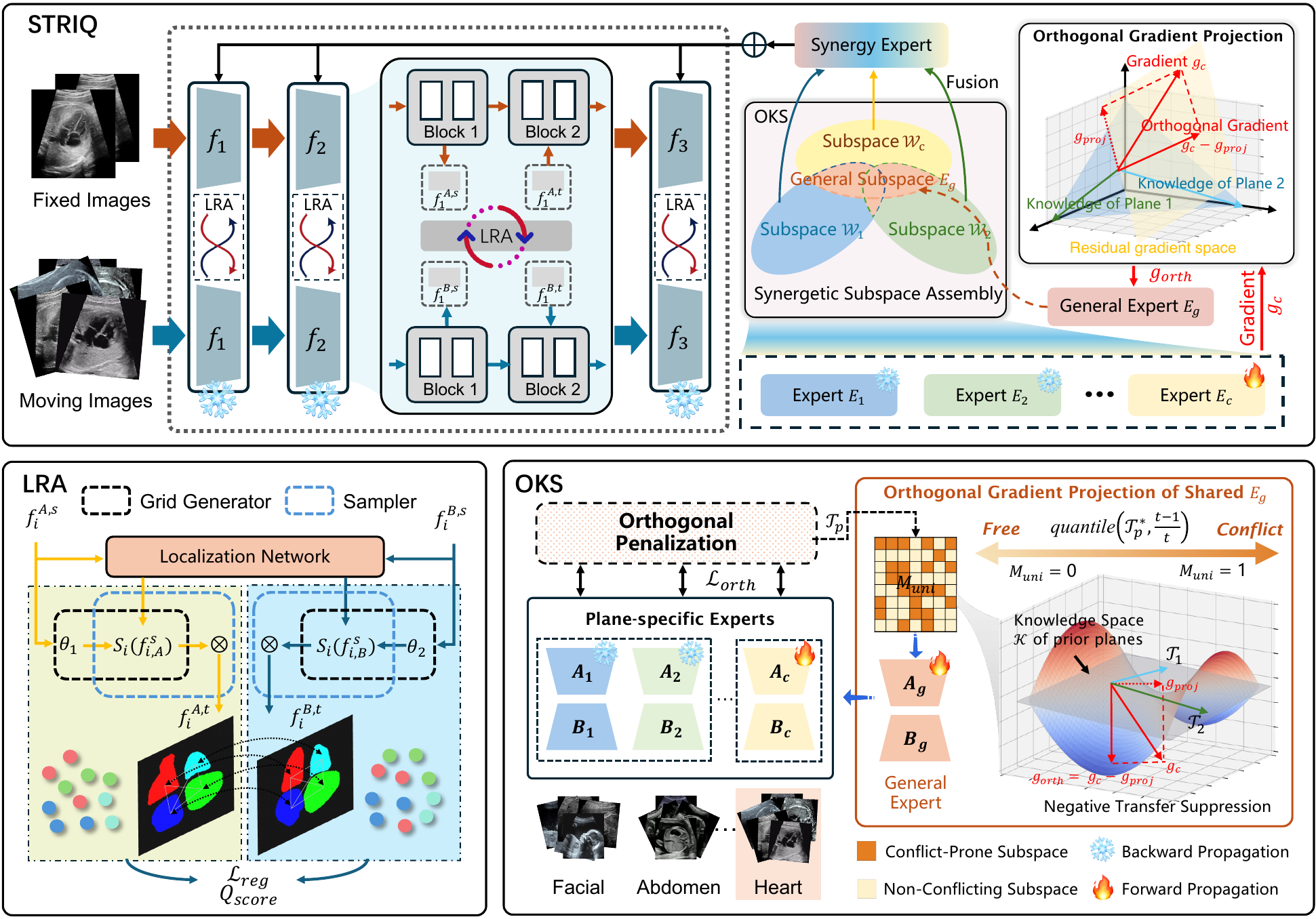}
\caption{Overview of \ours{}. A coupled feature extractor projects $(I_A,I_B)$ into hierarchical semantic embeddings $\{f_i\}_{i=1}^3$. The Latent Registration Aligner (LRA) progressively aligns these features via cascaded affine transformations, unifying feature-level registration and consistency-based quality evaluation (Sec.~\ref{sec:lra}, \ref{sec:inference}). The Orthogonal Knowledge Subspace (OKS) module decomposes the representations into orthogonal, plane-discriminative subspaces $\{\mathcal{S}_p\}_{p=1}^P$ to prevent inter-plane interference (Sec.~\ref{sec:oks}).}
\label{fig:framework}
\end{figure}

In this work, we present \ours, which addresses both challenges through two synergistic modules. The \textit{Latent Registration Aligner} (LRA) performs cascaded affine transformations in a hierarchical feature space, capturing correspondences from fine tissue boundaries to global organ positioning. The \textit{Orthogonal Knowledge Subspace} (OKS) module decomposes multi-plane representations into mutually orthogonal low-rank subspaces, enabling each anatomical plane to maintain a dedicated knowledge partition while facilitating controlled cross-plane knowledge sharing. During inference, OKS dynamically assembles a \textit{composite expert} by selectively fusing the most input-relevant subspace bases, achieving fine-grained plane discrimination without a conventional router.

Summary of our contributions: \textit{\textbf{(i)}} We propose \ours, a registration-driven QC framework that integrates topological invariance priors into feature-level similarity measurement, enabling annotation-free quality assessment applicable to both intra- and post-procedural scenarios. \textit{\textbf{(ii)}} We introduce the LRA module for hierarchical latent-space alignment and the OKS module for orthogonal plane-specific subspace decomposition, jointly resolving spatial deformation sensitivity and inter-plane knowledge conflict. \textit{\textbf{(iii)}} Evaluation on \textit{US4QA} (fetal) and \textit{CAMUS} (cardiac) datasets substantiates that \ours{} surpasses existing benchmarks in clinical alignment. Moreover, it maintains high-speed inference at ${\sim}5.6$\,ms/frame, satisfying real-time clinical requirements.


\section{Method}

\subsection{Problem Formulation and Anchor Selection}

Let $\mathcal{D} = \mathcal{D}_A \cup \mathcal{D}_B \cup \mathcal{D}_C$ partition an ultrasound dataset into a reference set $\mathcal{D}_A{=}\{I_A^j\}_{j=1}^{k_1}$ of variance-driven standard planes, a training set $\mathcal{D}_B{=}\{I_B^j\}_{j=1}^{k_2}$ with diverse quality levels ($k_1 \ll k_2$), and a query set $\mathcal{D}_C$ for evaluation. Each image is associated with an anatomical plane label $c \in \mathcal{C}$. We seek to learn a quality function $Q{:}\, \mathcal{I}{\to}[0,1]$ \emph{without manual quality annotations}, by optimising $\theta^* = \arg\min_\theta\,\mathbb{E}_{(I_A,I_B)\sim \mathcal{D}_A{\times}\mathcal{D}_B}[\mathcal{L}_\text{total}(\mathcal{N}_\theta(I_A),\,\mathcal{N}_\theta(I_B))]$, where $\mathcal{N}_\theta{:}\, \mathcal{I}{\to}\mathcal{F}$ denotes the hierarchical encoder equipped with LRA and OKS modules.

\vspace{1mm}
\noindent
\textit{Variance-Spectrum Reference Anchor Selection.}
For each plane $c\in\mathcal{C}$, we select $k_1$ reference images from the full corpus by minimizing intra-class feature embedding variance, following a variance-spectrum criterion. Using a pre-trained encoder $\mathcal{F}_{\text{pre}}$~\cite{lin2024beyond}, we compute embeddings for all candidate images and solve
$\mathcal{D}_A^{(c)} = \arg\min_{\mathcal{A}\subset \mathcal{X}_c,\,|\mathcal{A}|=k_1} \sum_{x\in\mathcal{A}} \sigma_{\mathcal{F}_{\text{pre}}}(x)^2,$
where $\sigma_{\mathcal{F}_{\text{pre}}}(x)^2$ quantifies the embedding variance of $x$ within plane $c$. This isolates high-confidence, morphologically consistent anchors that serve as device- and operator-invariant reference standards. The union $\mathcal{D}_A = \bigcup_{c\in\mathcal{C}}\mathcal{D}_A^{(c)}$ constitutes the complete reference library.

\subsection{Latent Registration Aligner (LRA)}
\label{sec:lra}
Given an image pair $(I_A, I_B)$, the Siamese encoder extracts hierarchical features via cascaded residual blocks $\{R_i\}_{i=1}^3$: $f_i^X = R_i \circ \cdots \circ R_1(X)$ for $X \in \{A, B\}$.
%
At each level $l$, a localisation network $\mathcal{L}_{loc}$ receives concatenated source features $[f_l^{A,s};\,f_l^{B,s}]$ and predicts an affine transformation $\Theta_l\in\text{SE}(2)$:
\begin{equation}
\small
\Theta_i = \mathcal{L}_\text{loc}^{(i)}\bigl([f_l^{A,s},\, f_l^{B,s}]\bigr) = \begin{bmatrix} a_{11} & a_{12} & t_x \\ a_{21} & a_{22} & t_y \end{bmatrix} \;\in\; \mathbb{R}^{2\times 3},
\end{equation}
where the source features are initialised as $f_1^{*,s}=f_1^{*}$ and cascaded via $f_l^{*,s}=f_{l-1}^{*,t}$ for $l>1$, with $f_l^{*,t}=\mathcal{G}(f_l^{*},\Theta_l)$ denoting bilinear grid sampling. 
This progressive alignment in feature space resolves multi-scale quality degradation, mitigating both global probe misalignments and local tissue distortions.


\vspace{2mm}
\noindent
\textit{Consistency Learning Objectives.}
To optimize the registration performance of the LRA, three complementary losses are employed to jointly enforce semantic, structural, and topological fidelity. First, semantic alignment $\mathcal{L}_{\texttt{sim}}$ facilitates the learning of discriminative representations on the unit hypersphere $\mathbb{S}^{d-1}$ by maximizing the cosine similarity between $\ell_2$-normalized features, defined as:
\begin{equation}
\small\small
-\sum_{l=1}^{3}\frac{1}{|\Omega_l|}\sum_{p\in\Omega_l}\langle \hat{f}_l^{A,t}(p),\,\hat{f}_l^{B,t}(p)\rangle, \hat{f}(p)=f(p)/\|f(p)\|_2.
\end{equation}
Then the local structural constraint $\mathcal{L}_{\texttt{NCC}}$ employs normalized cross-correlation (NCC) in feature space, which is inherently invariant to affine intensity transformations common in ultrasound, formulated as:
\begin{equation}
\small
-\sum_{l=1}^{3}\text{Cov}(f_l^{A,t}, f_l^{B,t} | \Omega_l) / (\sigma(f_l^{A,t} | \Omega_l) \cdot \sigma(f_l^{B,t} | \Omega_l)).
\end{equation}
Finally, a smoothness regularizer $\mathcal{L}_{\text{smooth}}$ is imposed on the displacement field $\psi: \Omega \rightarrow \mathbb{R}^2$ to promote inter-level geometric coherence and ensure anatomical plausibility. This constraint penalizes the squared Frobenius norm of the displacement Jacobian $\mathbf{J}_{\psi_l}$, formulated as:
$
\sum_{l=1}^{3} \int_{\Omega_l} |\mathbf{J}_{\psi_l}(p)|_F^2 , dp,
$
where $\Omega$ denotes the image domain. By enforcing this penalty, the displacement field remains within the Sobolev space $W^{1,2}(\Omega)$~\cite{balakrishnan2019voxelmorph}, thereby encouraging well-conditioned affine transitions across cascaded levels and ensuring structural continuity.

\subsection{Orthogonal Knowledge Subspace Module (OKS)}
\label{sec:oks}
When a unified registration network processes heterogeneous anatomical planes, \textit{negative transfer} arises from conflicting plane-specific distributions. OKS thus enforces mutually orthogonal plane-specific subspaces.


\vspace{1.5mm}
\noindent\textit{Plane-specific subspace decomposition.}
For each anatomical plane $c \in \mathcal{C}$, OKS maintains a low-rank expert
$E_c = B_c A_c$ with $A_c \in \mathbb{R}^{r \times d}$ and $B_c \in \mathbb{R}^{d \times r}$
($r \ll d$). The knowledge space is $\mathcal{S}_c = \text{span}\{s_c^1, \ldots, s_c^r\}$,
where each basis $s_c^j = \{A_c[j,:],\, B_c[:,j]\}$ encapsulates a distinct knowledge
dimension. To enforce inter-plane orthogonality, we minimise:
$
\mathcal{L}_\texttt{orth} = \frac{1}{|\mathcal{C}|(|\mathcal{C}|-1)}
\sum_{c \neq c'} \|A_c \cdot A_{c'}^\top\|_1,
\label{eq:orth}
$
which penalises representational overlap between any pair of subspaces
$(\mathcal{S}_c,\, \mathcal{S}_{c'})$.

\vspace{1.5mm}
\noindent\textit{Adaptive subspace selection.}
During inference, OKS computes the low-rank activation $z_c = A_c x \in \mathbb{R}^r$
for each expert $E_c$ and retains basis vectors whose activation exceeds threshold
$\epsilon$: $\mathcal{U}_c = \text{span}\{s_c^k \mid z_c[k] > \epsilon\}$.
To further suppress redundancy, only the top-$\kappa$ most activated bases are
preserved:
$\mathcal{V}_c = \text{span}\{u_c^l \mid l = \text{argtop}_\kappa(z_c)\}$,
where $\kappa = \min(|\mathcal{U}_c|,\, \lfloor r / |\mathcal{C}| \rfloor)$.

\vspace{1.5mm}
\noindent\textit{Task vector construction and conflict mask.}
After training on each plane $c$, we record a task vector
$\mathcal{T}_c = \{T_c^A,\, T_c^B\}$, where $T_c^A = A_c^{ft} - A_c^{pre}$
and $T_c^B = B_c^{ft} - B_c^{pre}$ capture the net parameter shift.
A binary mask $\mathcal{M}_c = \{M_c^A,\, M_c^B\}$ then identifies the most
critical parameters of $E_c$:
\begin{equation}
\small
M_c^*[i,j] =
\begin{cases}
1, & T_c^*[i,j] > \mathrm{quantile}(T_c^*,\;\tfrac{|\mathcal{C}|-1}{|\mathcal{C}|}), \\
0, & \text{otherwise},
\end{cases}
\quad * \in \{A,B\}.
\end{equation}
The union mask $\mathcal{M}^{\text{uni}} = \bigcup_{c \in \mathcal{C}} \mathcal{M}_c$
partitions subspace parameters into conflict-prone ($\mathcal{M}^{\text{uni}}\!=\!1$)
and conflict-free regions ($\mathcal{M}^{\text{uni}}\!=\!0$), as illustrated in Fig.~\ref{fig:framework}.

\vspace{1.5mm}
\noindent\textit{Shared knowledge with orthogonal projection.}
OKS maintains a general expert $E_g = B_g A_g$ to capture cross-plane shared
representations. The prior knowledge space
$\mathcal{K} \in \mathbb{R}^{(|\mathcal{C}|-1)\times(r \cdot d)}$ is obtained by
normalising and stacking previously seen task vectors. When learning from plane
$c$, the gradient $g_c$ is projected onto the orthogonal complement of $\mathcal{K}$:
$g_{\text{orth}} = g_c - \mathcal{K}^\top \mathcal{K}\, g_c$.
The general expert is then updated as:
$
E_g^* \leftarrow E_g^* - \eta\,
\bigl(\mathcal{M}^{\text{uni},*} \odot g_{\text{orth}}
+ (1 - \mathcal{M}^{\text{uni},*}) \odot g_c\bigr),
\quad * \in \{A_g, B_g\},
$
where $\eta$ is the learning rate and $\odot$ denotes element-wise product.
This selectively suppresses interference in conflict-prone subspaces while
preserving direct gradient flow elsewhere.

\vspace{1.5mm}
\noindent\textit{Synergy Expert assembly.}
During inference, relevant bases are also extracted from $E_g$ via activation
thresholding:
$\mathcal{W}_g = \text{span}\{s_g^k \mid z_g[k] > \gamma,\; z_g = A_g x\}$,
yielding $\{A^g_{\mathcal{W}}, B^g_{\mathcal{W}}\}$.
All selected plane-specific and general subspace matrices are concatenated to
form the Synergy Expert $\mathcal{E} = \{A^\mathcal{E}, B^\mathcal{E}\}$:
\begin{equation}
\small
A^\mathcal{E} =
\bigl[A_{\mathcal{V}_1}^\top,\ldots,A_{\mathcal{V}_{|\mathcal{C}|}}^\top,
(A^g_{\mathcal{W}})^\top\bigr]^\top,\quad
B^\mathcal{E} =
\bigl[B_{\mathcal{V}_1},\ldots,B_{\mathcal{V}_{|\mathcal{C}|}},
B^g_{\mathcal{W}}\bigr],
\end{equation}
and the final feature output is computed as:
$\hat{f} = W_0 x + \tfrac{\alpha}{r}\, B^\mathcal{E} A^\mathcal{E} x$,
where $W_0$ denotes the frozen backbone weights and $\alpha$ modulates the
update magnitude. This activation-driven assembly enables fine-grained plane discrimination while preserving complementary cross-plane knowledge.

\subsection{Training and Quality Score Inference}
\label{sec:inference}
The total training objective is:
$
\mathcal{L}_{\text{total}} = \mathcal{L}_{\texttt{reg}}+ \lambda\,\mathcal{L}_{\texttt{orth}}$,
where $\mathcal{L}_{\texttt{reg}} = \mathcal{L}_{\texttt{sim}} + \mathcal{L}_{\texttt{NCC}} + \mathcal{L}_{\texttt{smooth}}$ subsumes the three registration consistency terms from Sec.~\ref{sec:lra}, and $\lambda$ controls the trade-off between subspace separation and registration fidelity. During inference, the quality score for a query image $I_C \in \mathcal{D}_C$ is:
\begin{equation}
\small
Q(I_C) = 1 - \frac{1}{k_1}\sum_{j=1}^{k_1}\sum_{m\in\mathcal{M}}w_m\,\phi\!\bigl(\mathcal{L}_m(I_C, I_A^{(j)})\bigr),
\end{equation}
where $\mathcal{M}=\{\texttt{sim},\texttt{NCC},\texttt{smooth}\}$ and $\phi(\cdot)$ applies min-max normalisation over accumulated statistics. An image is deemed clinically acceptable when $Q(I_C)>\tau$, with the threshold $\tau$ adjustable per clinical protocol that directly supports both intra-scan real-time gating and post-hoc batch QC.



\section{Experiments}
\label{sec:experiments}
\vspace{-1mm}
\subsection{Experimental Setup}

\paragraph{Implementation Details.}
\ours{} is implemented in PyTorch with a ResNet-18 backbone $\mathcal{N}_\theta$ initialised from ImageNet pre-trained weights (frozen during training). The LRA localisation networks and OKS expert matrices $\{A_c,B_c\}_{c\in\mathcal{C}}$  with the general expert $\{A_g,B_g\}$ are optimised via Adam (lr $= 1{\times}10^{-4}$, 500 epochs, batch size 64) on an NVIDIA RTX 4090 (24 GB). The OKS low-rank dimension is set to $r{=}16$, the subspace orthogonality weight to $\lambda{=}0.5$, and the activation thresholds to $\epsilon{=}\gamma{=}0.1$. The quality score weights $w_m$ are all set to 1; the acceptance threshold is $\tau{=}0.5$. We employ 5-fold cross-validation on $\mathcal{D}_B$, with $\mathcal{D}_C$ for final evaluation. Data augmentation includes contrast adjustment, rotation (${\pm}20°$), translation (${\pm}20\%$), scaling ($0.8$--$1.2{\times}$), and Gaussian noise injection.

\vspace{-2mm}
\paragraph{Baselines, Datasets, and Evaluation Protocol.}
We evaluate \ours{} on \textit{US4QA} (30,757 fetal ultrasound images across four planes: 4CH, abdomen, kidneys, face; 324 examinations) and the public \textit{CAMUS} dataset~\cite{leclerc2019deep} (A2C/A4C cardiac views). For \textit{US4QA}, $k_1{=}20$ images per plane are selected as reference anchors ($\mathcal{D}_A$) via the variance-spectrum criterion; the remainder is split 9:1 into $\mathcal{D}_B$ and $\mathcal{D}_C$, with ground-truth scores averaged from six sonographers on $[0,1]$. For \textit{CAMUS}, ordinal grades (\textit{Good/Medium/Poor}) are treated as a ranked sequence. Performance is reported via SRCC~\cite{sedgwick2014spearman} and PLCC~\cite{sedgwick2012pearson}. Baselines span four categories: (\textit{i})~traditional NR-IQA methods (BRISQUE~\cite{mittal2012no}, MSSIM~\cite{wang2004image}); (\textit{ii})~structure-detection confidence methods (ARVBNet~\cite{dong2019generic} for 4CH, FUIQA~\cite{wu2017fuiqa} for abdomen, MF R-CNN~\cite{lin2019multi} for face and kidneys), collectively referred to as Weighted Score; (\textit{iii})~unsupervised IQA methods (ARNIQA~\cite{agnolucci2024arniqa}, SFD-IQA~\cite{dong2025sfd}); and (\textit{iv})~the pseudo-label-based SOTA approach CRL-UIQA~\cite{guo2024unsupervised}.

\begin{table*}[t]
    \centering
    \caption{%
        Quantitative comparison (SRCC) of \ours{} against competing methods on the \textit{US4QA} and \textit{CAMUS}\,\textsuperscript{\dag} test sets.
        \textbf{Bold}: best; \underline{underlined}: second-best.
    }
    \label{tab:main_comparison}
    \renewcommand{\arraystretch}{1}
    \resizebox{0.9\linewidth}{!}{%
    \begin{tabular}{p{1.8cm} l @{\hspace{6pt}} c @{\hspace{6pt}} c @{\hspace{6pt}} c @{\hspace{6pt}} c @{\hspace{6pt}} c  @{\hspace{6pt}}c @{\hspace{6pt}} c}
    \toprule[1.5pt]
    \textbf{Dataset}
        & \textbf{Plane}
        & \textbf{\makecell{BRISQUE\\\cite{mittal2012no}}}
        & \textbf{\makecell{MSSIM\\\cite{wang2004image}}}
        & \textbf{\makecell{ARNIQA\\\cite{agnolucci2024arniqa}}}
        & \textbf{\makecell{SFD-IQA\\\cite{dong2025sfd}}}
        & \textbf{\makecell{Weight\\Score}}
        & \textbf{\makecell{CRL-UIQA\\\cite{guo2024unsupervised}}}
        & \textbf{\makecell{\ours{}\\(Ours)}} \\
    \midrule[1.2pt]

    \multirow{5}{*}{\textit{US4QA}}
        & \textbf{Abdomen}
          & 0.389 & 0.217 & 0.563 & 0.589 & 0.738 & \underline{0.801} & \textbf{0.847} \\
        & \textbf{4CH}
          & 0.407 & 0.405 & 0.494 & 0.477 & 0.679 & \underline{0.825} & \textbf{0.883} \\
        & \textbf{Kidney}
          & 0.449 & 0.401 & 0.523 & 0.549 & 0.724 & \underline{0.731} & \textbf{0.819} \\
        & \textbf{Face}
          & 0.488 & 0.469 & 0.571 & 0.533 & 0.692 & \underline{0.759} & \textbf{0.831} \\
    \cmidrule(lr){2-9}
        & \textit{Average}
          & 0.433 & 0.373 & 0.538 & 0.537 & 0.708 & \underline{0.779} & \textbf{0.845} \\

    \midrule[1.2pt]

    \multirow{3}{*}{\textit{CAMUS}\textsuperscript{\dag}}
        & \textbf{A2C}
          & 0.341 & 0.298 & 0.512 & 0.527 & {---} & \underline{0.683} & \textbf{0.762} \\
        & \textbf{A4C}
          & 0.367 & 0.312 & 0.538 & 0.541 & {---} & \underline{0.701} & \textbf{0.779} \\
    \cmidrule(lr){2-9}
        & \textit{Average}
          & 0.354 & 0.305 & 0.525 & 0.534 & {---} & \underline{0.692} & \textbf{0.771} \\

    \bottomrule[1.5pt]
    \end{tabular}%
    }
\end{table*}

\begin{figure*}[t]
\centering
\includegraphics[width=\linewidth]{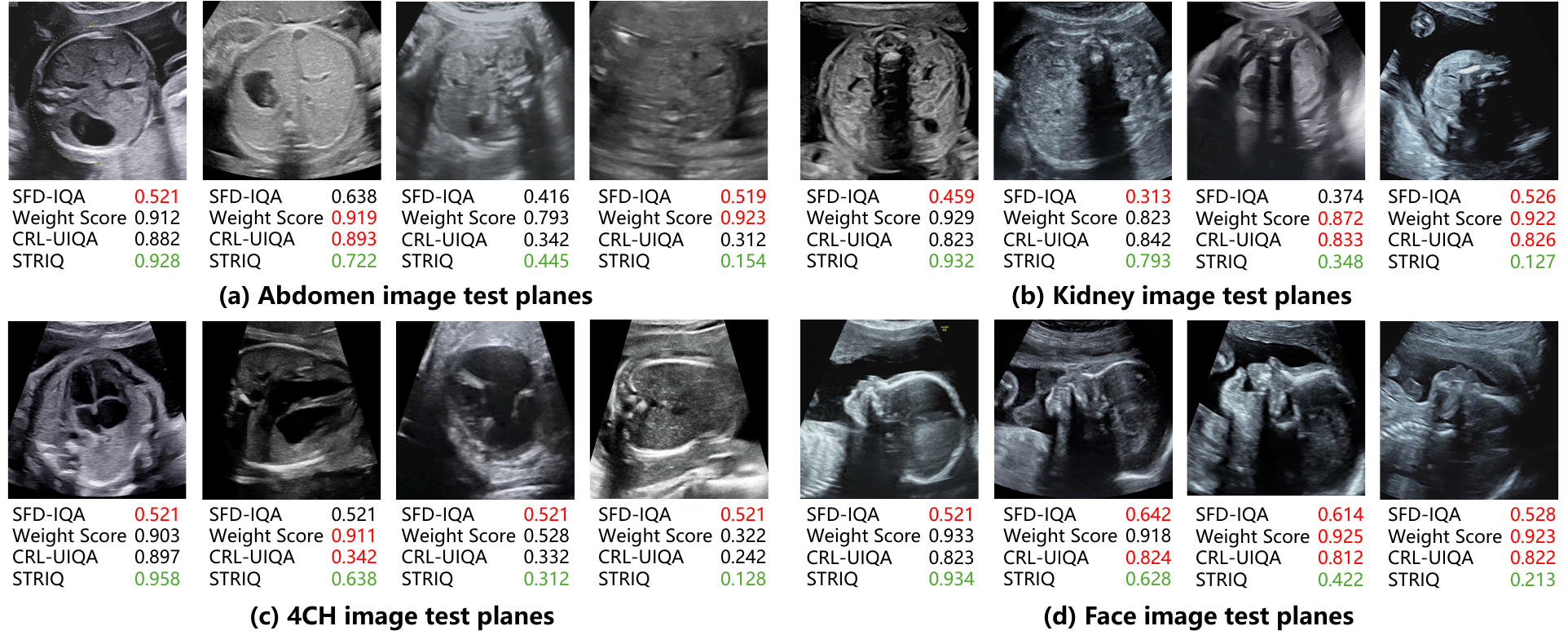}
\caption{Qualitative comparison on the US planes. Each group is sorted from high to low clinical quality (left to right). \textcolor[HTML]{4EA62F}{Green}: accurate assessments; \textcolor{red}{red}: prediction errors.}
\label{fig:qualitative}
\end{figure*}

\subsection{Comparison with State-of-the-Art Methods}
As shown in Table~\ref{tab:main_comparison}, \ours{} consistently attains the highest SRCC across all planes. On \textit{US4QA}, \ours{} surpasses CRL-UIQA by an average of $\mathbf{+6.6\%}$ (avg.\ $\mathbf{0.845}$ vs.\ $0.779$), with \textit{\textbf{the largest margins on Kidney and Face}} where inter-plane confusion is most pronounced and pseudo-label methods suffer from deformation-induced embedding collapse. On \textit{CAMUS}, jointly trained with \textit{US4QA}, \ours{} further outperforms CRL-UIQA by $\mathbf{+7.8\%}$ (avg.\ $\mathbf{0.771}$), confirming cross-anatomy robustness. The consistent collapse of generic NR-IQA methods (avg.\ ${\approx}\,0.538$) across both datasets validates the \textit{\textbf{necessity of topological invariance priors}} for ultrasound QC.
Fig.~\ref{fig:qualitative} provides qualitative corroboration. \ours{} assigns monotonically decreasing scores as anatomical structures become incomplete and artifacts accumulate, whereas CRL-UIQA produces conspicuous mispredictions on Kidney and Face planes where embedding-space proximity fails under geometric deformation. By grounding quality estimation in subspace-guided registrability, \ours{} avoids this failure mode.
%
%
%
%
%

\begin{table*}[t]
    \centering
    \caption{%
        Ablation analysis of \ours{} evaluated by SRCC on the $\mathcal{D}_C$ test set.
        \textit{Left}: architectural component ablation.
        \textit{Right}: loss function contribution analysis.
        Statistical significance against the full model is assessed via paired $t$-tests:
        $^{\dagger}$~$p < 0.05$;\;$^{\ddagger}$~$p < 0.01$.
    }
    \label{tab:ablation_combined}
    \renewcommand{\arraystretch}{0.9}
    \begin{minipage}[t]{0.492\linewidth}
        \centering
        \resizebox{\linewidth}{!}{%
        \begin{tabular}{p{3.7cm} >{\centering\arraybackslash}p{1cm}>{\centering\arraybackslash}p{1cm}>{\centering\arraybackslash}p{1cm}>{\centering\arraybackslash}p{1cm}}
        \toprule[1.3pt]
        \textbf{Configuration} & \textbf{Abd.} & \textbf{4CH} & \textbf{Kid.} & \textbf{Avg.} \\
        \midrule[1.0pt]
        \textbf{\ours{} (Full)}
            & \textbf{0.847} & \textbf{0.883} & \textbf{0.819} & \textbf{0.845} \\
        \midrule
        \multicolumn{5}{l}{\textit{Module Ablations}} \\[1pt]
        \;\textit{w/o} OKS module
            & 0.789\rlap{$^{\ddagger}$} & 0.821\rlap{$^{\ddagger}$} & 0.768\rlap{$^{\ddagger}$} & 0.793\rlap{$^{\ddagger}$} \\
        \;\textit{w/o} Orthogonality            & 0.801\rlap{$^{\dagger}$}  & 0.838\rlap{$^{\dagger}$}  & 0.779\rlap{$^{\dagger}$}  & 0.806\rlap{$^{\dagger}$}  \\
        \;\textit{w/o} Adaptive routing
            & 0.812\rlap{$^{\dagger}$}  & 0.851\rlap{$^{\dagger}$}  & 0.791\rlap{$^{\dagger}$}  & 0.818\rlap{$^{\dagger}$}  \\
        \;\textit{w/o} Hierarchical LRA
            & 0.731\rlap{$^{\ddagger}$} & 0.792\rlap{$^{\ddagger}$} & 0.724\rlap{$^{\ddagger}$} & 0.749\rlap{$^{\ddagger}$} \\
        \bottomrule[1.3pt]
        \end{tabular}%
        }
    \end{minipage}%
    \hfill
    \renewcommand{\arraystretch}{0.92}
    \begin{minipage}[t]{0.465\linewidth}
        \centering
        \resizebox{\linewidth}{!}{%
        \begin{tabular}{>{\centering\arraybackslash}p{0.7cm}>{\centering\arraybackslash}p{0.7cm} >{\centering\arraybackslash}p{0.7cm} >{\centering\arraybackslash}p{0.7cm}>{\centering\arraybackslash}p{1cm} >{\centering\arraybackslash}p{1cm} >{\centering\arraybackslash}p{1cm} >{\centering\arraybackslash}p{1cm} }
        \toprule[1.3pt]
        $\mathcal{L}_{\texttt{sim}}$ & $\mathcal{L}_{\texttt{NCC}}$ & $\mathcal{L}_{\texttt{smo}}$ & $\mathcal{L}_{\texttt{orth}}$
            & \textbf{Abd.} & \textbf{4CH} & \textbf{Kid.} & \textbf{Avg.} \\
        \midrule[1.0pt]
        \checkmark & \checkmark & \checkmark & \checkmark
            & \textbf{0.847} & \textbf{0.883} & \textbf{0.819} & \textbf{0.845} \\
        \checkmark & \checkmark & \checkmark &
            & 0.801\rlap{$^{\dagger}$} & 0.838\rlap{$^{\dagger}$} & 0.779\rlap{$^{\dagger}$} & 0.806\rlap{$^{\dagger}$} \\
        \checkmark & \checkmark &            & \checkmark
            & 0.815\rlap{$^{\dagger}$} & 0.849\rlap{$^{\dagger}$} & 0.798\rlap{$^{\dagger}$} & 0.821\rlap{$^{\dagger}$} \\
            
            \checkmark &            &            & 
            & 0.712\rlap{$^{\ddagger}$} & 0.705\rlap{$^{\ddagger}$} & 0.719\rlap{$^{\ddagger}$} & 0.712\rlap{$^{\ddagger}$} \\
        \checkmark &            & \checkmark & \checkmark
            & 0.808\rlap{$^{\dagger}$} & 0.831\rlap{$^{\dagger}$} & 0.795\rlap{$^{\dagger}$} & 0.811\rlap{$^{\dagger}$} \\
             & \checkmark & \checkmark & \checkmark
            & 0.682\rlap{$^{\ddagger}$} & 0.671\rlap{$^{\ddagger}$} & 0.701\rlap{$^{\ddagger}$} & 0.685\rlap{$^{\ddagger}$} \\
        \bottomrule[1.3pt]
        \end{tabular}%
        }
    \end{minipage}
\end{table*}

\begin{figure}[t]
    \centering
    \includegraphics[width=\linewidth]{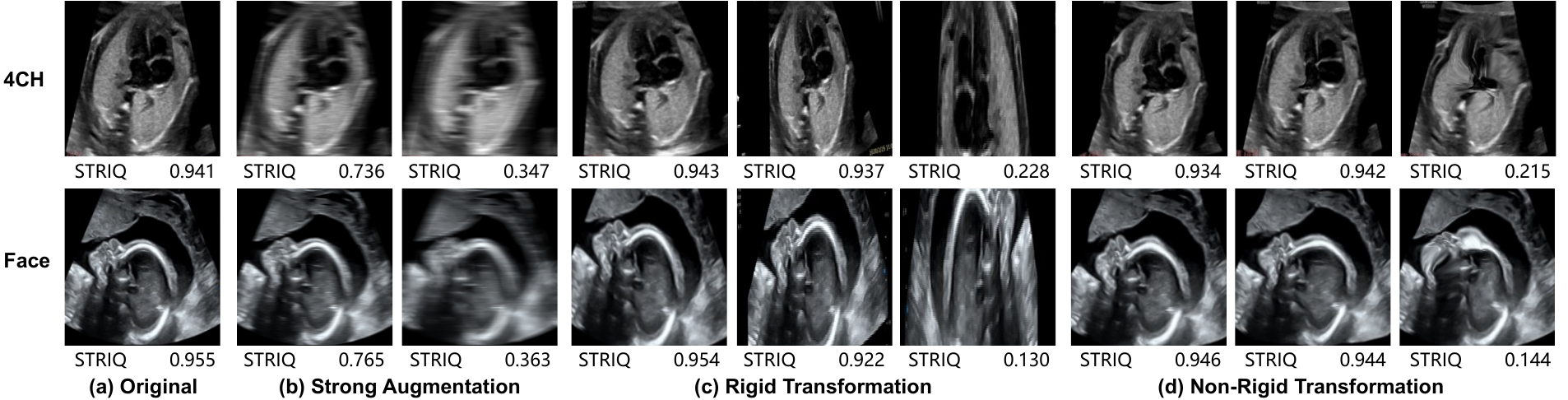}
\caption{Quality score response of \ours{} under rigid and non-rigid deformations of increasing severity, ranging from minor perturbations that preserve structural integrity to excessive deformations inducing anatomical absence or atypical formation.}

    \label{fig:deformation_robustness}
\end{figure}

\vspace{-1mm}
\subsection{Comprehensive Ablation Analysis}

As shown in Table~\ref{tab:ablation_combined}, \textit{\textbf{OKS removal}} induces the sharpest degradation ($0.793$ avg., $p{<}0.01$), as the backbone conflates heterogeneous plane representations without dedicated subspaces; ablating orthogonality alone yields $0.806$ ($p{<}0.05$), with 4CH most affected ($0.838$ vs.\ $0.883$). \textit{\textbf{Hierarchical LRA}} is equally critical: single-stage alignment collapses to $0.749$ ($p{<}0.01$). On the loss side, \textit{\textbf{$\mathcal{L}_{\texttt{sim}}$ is dominant}} (removal: $0.685$, $p{<}0.01$), yet $\mathcal{L}_{\texttt{sim}}$ alone yields only $0.712$, confirming that $\mathcal{L}_{\texttt{NCC}}$, $\mathcal{L}_{\texttt{smooth}}$, and $\mathcal{L}_{\texttt{orth}}$ jointly supply structural correspondence, diffeomorphic plausibility, and inter-plane disentanglement. To sum up, each component and loss term contributes meaningfully to the final performance.

Fig.~\ref{fig:deformation_robustness} evaluates \ours{} under progressive rigid and non-rigid deformations. Minor deformations preserve slice characteristics and yield high scores, whereas severe deformations causing structural absence or atypical formation result in proportionally decreased scores. This \textit{\textbf{consistent score-deformation correspondence}} confirms that \ours{} precisely captures clinically dynamic deformations and geometric-topological integrity of anatomical structures.

\section{Conclusion}
In this work, we propose \ours{}, which reformulates ultrasound quality control as subspace-guided consistency measurement without manual annotations. By introducing LRA for hierarchical latent-space alignment and OKS for orthogonal plane-specific subspace decomposition, \ours{} jointly addresses spatial deformation sensitivity and inter-plane knowledge conflict across multiple diagnostic planes, demonstrating consistent superiority over state-of-the-art methods with robust quality tracking under intra- and post-procedural clinical deployment.

\begin{credits}
\subsubsection{Acknowledgments.} This research was partially supported by the National Major Scientific Instrument Development Project of the National Natural Science Foundation of China (Grant No. 62227808), the National Natural Science Foundation of China (Grants No. 62472157 and No. 62272159), the Key Research and Development Program of Ningxia Hui Autonomous Region (Grant No. 2024BEG02018), the Natural Science Foundation of Hunan Province (Grant No. 2024JJ5089), the Guangdong Basic and Applied Basic Research Foundation (Grant No. 2025A1515011404), and the Hunan Provincial Graduate Research Innovation Project (Grant No. CX20250587).
\subsubsection{Disclosure of Interests.}The authors have no competing interests to declare that are relevant to the content of this article.
\end{credits}


\bibliographystyle{splncs04}
\bibliography{reference}

\clearpage
\newpage

\setcounter{subsection}{0}
\renewcommand{\thesubsection}{A.\arabic{subsection}}


\setcounter{subsection}{0}
\renewcommand{\thesubsection}{A.\arabic{subsection}}

\vspace{-2mm}
\begin{table}[t]
\centering
\caption{Transformation modes supported by LRA.}
\begin{tabular}{cc|cc}
\toprule
\textbf{Mode} & \textbf{Matrix Formula} & \textbf{Mode} & \textbf{Matrix Formula} \\
\hline
\textbf{affine} & $\theta = \begin{bmatrix} a & b & tx \\ c & d & ty \end{bmatrix}$ & \textbf{shear} & $\theta = \begin{bmatrix} 1 & \lambda_x & 0 \\ \lambda_y & 1 & 0 \end{bmatrix}$ \\
\textbf{translation} & $\theta = \begin{bmatrix} 1 & 0 & tx \\ 0 & 1 & ty \end{bmatrix}$ & \textbf{rotation\_scale} & $\theta = \begin{bmatrix} s_x \cos\theta & -\sin\theta & 0 \\ \sin\theta & s_y \cos\theta & 0 \end{bmatrix}$ \\
\textbf{rotation} & $\theta = \begin{bmatrix} \cos\theta & -\sin\theta & 0 \\ \sin\theta & \cos\theta & 0 \end{bmatrix}$ & \textbf{translation\_scale} & $\theta = \begin{bmatrix} s_x & 0 & tx \\ 0 & s_y & ty \end{bmatrix}$ \\
\textbf{scale} & $\theta = \begin{bmatrix} s_x & 0 & 0 \\ 0 & s_y & 0 \end{bmatrix}$ & \textbf{rotation\_translation} & $\theta = \begin{bmatrix} \cos\theta & -\sin\theta & tx \\ \sin\theta & \cos\theta & ty \end{bmatrix}$ \\
\bottomrule
\end{tabular}
\label{tab:transform}
\end{table}

\begin{table}[t]
\centering
\caption{Hyperparameter sensitivity (SRCC avg.). Default \textbf{bolded}; $\uparrow$: peak.}
\label{tab:hyperparam_ext}
\small
\setlength{\tabcolsep}{3pt}
\resizebox{0.95\textwidth}{!}{%
\begin{tabular}{cc|cc|cc|cc|cc|cc}
\toprule
$r$ & SRCC & $\lambda$ & SRCC & $\epsilon{=}\gamma$ & SRCC & $k_1$ & SRCC & $w_{\texttt{sim}}{:}w_{\texttt{NCC}}{:}w_{\texttt{smo}}$ & SRCC & $\tau$ & F1 \\
\cmidrule{1-2}\cmidrule{3-4}\cmidrule{5-6}\cmidrule{7-8}\cmidrule{9-10}\cmidrule{11-12}
4  & 0.791 & 0.1 & 0.821 & 0.0 & 0.823 & 10  & 0.819 & 1:0:0 & 0.712 & 0.3 & 0.781 \\
8  & 0.818 & 0.3 & 0.836 & \textbf{0.1} & \textbf{0.845}$\uparrow$ & 15  & 0.831 & 1:1:0 & 0.811 & 0.4 & 0.823 \\
\textbf{16} & \textbf{0.845}$\uparrow$ & \textbf{0.5} & \textbf{0.845}$\uparrow$ & 0.2 & 0.839 & \textbf{20} & \textbf{0.845}$\uparrow$ & \textbf{1:1:1} & \textbf{0.845}$\uparrow$ & \textbf{0.5} & \textbf{0.861}$\uparrow$ \\
32 & 0.841 & 0.8 & 0.831 & 0.4 & 0.831 & 25 & 0.843 & 2:1:1 & 0.840 & 0.6 & 0.849 \\
64 & 0.833 & 1.0 & 0.819 & 0.6 & 0.812 & 30 & 0.841 & 1:2:1 & 0.837 & 0.7 & 0.832 \\
\bottomrule
\end{tabular}
}
\end{table}

\subsection{Supplementary Experimental Protocol}


\inlinesec{Scoring and Latency.}
The min-max normalisation $\phi(\cdot)$ is computed per-plane over $\mathcal{D}_B$ and \textit{fixed} at inference to prevent distribution shift. Score weights are uniform ($w_m{=}1$), as no single term dominates the quality manifold; $\tau{=}0.5$ maximises accept/reject F1. Anchor features are \textit{\textbf{pre-computed and cached}}: with $k_1{=}20$, inference takes ${\sim}5.6$\,ms/frame (${\sim}180$ FPS) on an RTX 4090.

\inlinesec{Affine Transformation.}
The LRA predicts affine matrices $\Theta_i \in \mathbb{R}^{2 \times 3}$. The cascaded architecture applies distinct affine transformations per feature level; the smoothness penalty regularises \textit{\textbf{inter-level transition continuity}} rather than enforcing diffeomorphism. Table~\ref{tab:transform} enumerates the eight supported modes.

\begin{table}[t]
    \centering
    \caption{%
        \textit{Left}: Anchor selection strategy.
        \textit{Right}: $\mathcal{L}_{\texttt{orth}}$ variant (SRCC avg.\ on $\mathcal{D}_C$).
    }
    \vspace{-2mm}
    \label{tab:runtime_orth}
    \renewcommand{\arraystretch}{0.95}
    \begin{minipage}[t]{0.46\linewidth}
        \centering
        \resizebox{\linewidth}{!}{%
        \begin{tabular}{l c c c c}
        \toprule[1.3pt]
        \textbf{Strategy} & \textbf{Abd.} & \textbf{4CH} & \textbf{Kid.} & \textbf{Avg.} \\
        \midrule[1.0pt]
        Random & 0.798 & 0.832 & 0.771 & 0.798 \\
        K-medoids & 0.831 & 0.862 & 0.801 & 0.827 \\
        K-center greedy & 0.825 & 0.855 & 0.796 & 0.821 \\
        \textbf{Var.-spectrum} & \textbf{0.847} & \textbf{0.883} & \textbf{0.819} & \textbf{0.845} \\
        \bottomrule[1.3pt]
        \end{tabular}}
    \end{minipage}
    \hfill
    \begin{minipage}[t]{0.48\linewidth}
        \centering
        \resizebox{\linewidth}{!}{%
        \begin{tabular}{l c c c c}
        \toprule[1.3pt]
        \textbf{Variant} & \textbf{Abd.} & \textbf{4CH} & \textbf{Kid.} & \textbf{Avg.} \\
        \midrule[1.0pt]
        $\|A_c A_{c'}^\top\|_F$  & 0.841 & 0.876 & 0.815 & 0.839 \\
        $\|\cdot\|_1$ on $A{+}B$  & 0.843 & 0.879 & 0.817 & 0.841 \\
        $\|\cdot\|_F$ on $A{+}B$  & 0.838 & 0.871 & 0.811 & 0.835 \\
        $\|A_c A_{c'}^\top\|_1$ (\textbf{default}) & \textbf{0.847} & \textbf{0.883} & \textbf{0.819} & \textbf{0.845} \\
        \bottomrule[1.3pt]
        \end{tabular}}
    \end{minipage}
\end{table}

\begin{figure}[htpb]
    \centering
    \includegraphics[width=1\linewidth]{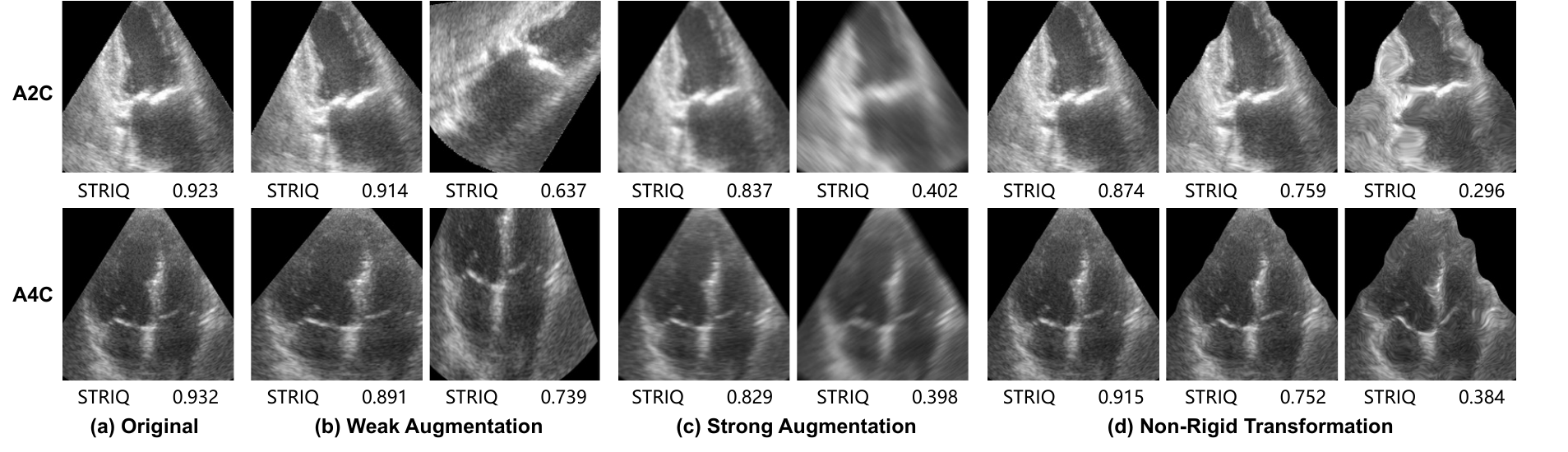}
    \caption{Quality score response of \ours{} on CAMUS A2C/A4C planes under (a) original, (b) weak augmentation, (c) strong augmentation, and (d) non-rigid transformation with increasing severity (left to right within each group).}
    \vspace{-2mm}
    \label{fig:camus_vis}
\end{figure}

\begin{figure}[t]
    \centering
    \includegraphics[width=\linewidth]{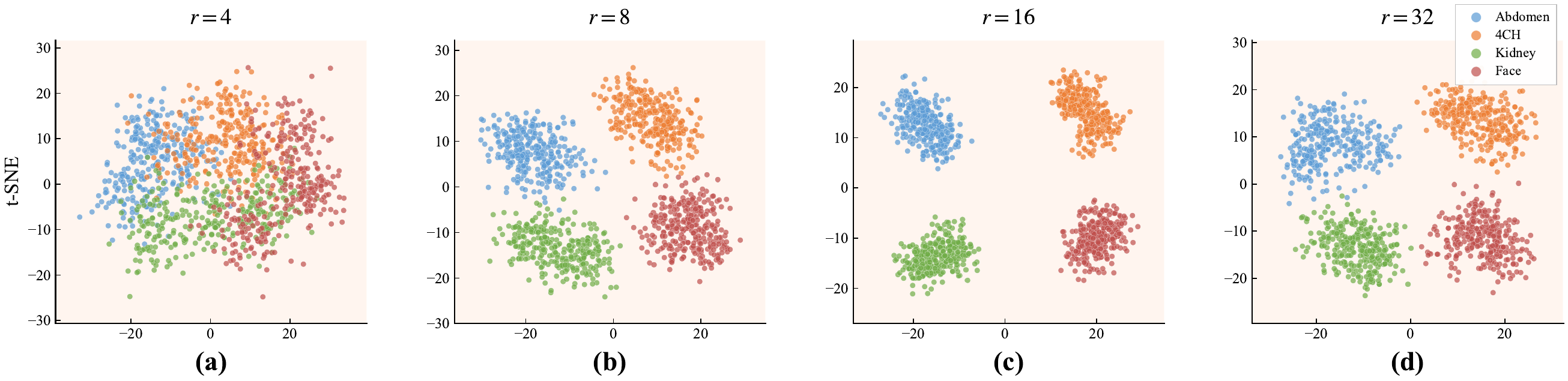}
    \caption{OKS subspace embeddings under varying ranks. (a)~$r{=}4$: overlap. (b)~$r{=}8$: partial separation. (c)~$r{=}16$: \textit{\textbf{optimal compact clusters}}. (d)~$r{=}32$: mild redundancy.}
    \vspace{-1mm}
    \label{fig:oks_tsne}
\end{figure}

\subsection{Anchor Strategy, $\mathcal{L}_{\texttt{orth}}$ Variants, and Hyperparam. Sensitivity}
Table~\ref{tab:runtime_orth} (left) shows variance-spectrum selection outperforms k-medoids ($+1.8\%$), while random selection degrades to $0.798$, confirming the criticality of anchor curation. Table~\ref{tab:runtime_orth} (right) shows the default $\ell_1$ norm on $A_c$ achieves the best subspace separation ($0.845$), as $\ell_1$ promotes sparser cross-Gram entries than Frobenius alternatives. Extending the penalty to both projection matrices ($\|\cdot\|$ on $A{+}B$, \textit{i.e.}\ $\|A_c A_{c'}^\top\|{+}\|B_c^\top B_{c'}\|$) yields no gain, indicating that penalising the down-projection cross-Gram alone suffices for subspace disentanglement. Table~\ref{tab:hyperparam_ext} reports sensitivity across six hyperparameters: \textit{\textbf{OKS rank}} $r{=}16$ balances expressiveness and separability; \textit{\textbf{anchor size}} $k_1{=}20$ suffices to capture each plane's morphological distribution, with no gain beyond this; \textit{\textbf{threshold}} $\tau{=}0.5$ maximizes F1, exhibiting robustness across the $[0.4, 0.6]$ interval.

\subsection{Other Results and Subspace Analysis}
\label{sec:supp_vis}

\inlinesec{CAMUS Results.}
Qualitative results in Fig.~\ref{fig:camus_vis} show that \ours{} scores exhibit graceful decay under mild noise yet decrease proportionally to structural loss under severe deformation. This confirms that subspace-guided registrability \textit{\textbf{captures fundamental ultrasound geometry}}, effectively accommodating diverse cardiac morphologies within the jointly trained framework.

\inlinesec{OKS Subspace Separation.}
The impact of OKS rank on feature distribution is visualized via t-SNE in Fig.~\ref{fig:oks_tsne}. While a low rank ($r{=}4$) results in significant inter-plane entanglement, $r{=}16$ yields \textit{\textbf{distinct, compact clusters}} that align with the observed SRCC peak. Increasing the rank to $r{=}32$ introduces marginal redundancy without further enhancing separation, reinforcing $r{=}16$ as the optimal balance between discriminability and manifold compactness.

\end{document}